\documentclass[final,5p,times,twocolumn,number]{elsarticle}

\global\bibsep=0pt 

\usepackage{amssymb}
\usepackage{algorithm}
\usepackage{algpseudocode}
\usepackage{graphicx}
\newcommand{\ie}{{\emph{i.e.}}, }
\newcommand{\etal}{{\it {et al}}. }

\usepackage{subfigure}
\usepackage{amsmath}

\usepackage{xcolor}
\usepackage{tabularx} 
\usepackage{multirow}
\usepackage{multicol}
\usepackage{booktabs} 
\usepackage{pbox}
\usepackage{amsfonts}
\usepackage{graphicx}
{\begin{list}               
    {$\bullet$ \hfill}{
        \setlength{\leftmargin}{\parindent}
        \setlength{\parsep}{0.04\baselineskip}
        \setlength{\itemsep}{0.5\parsep}
        \setlength{\labelwidth}{\leftmargin}
        \setlength{\labelsep}{0em}}
    }
{\end{list}}

\providecommand{\eref}[1]{Eq. \eqref{#1}}  
\providecommand{\cref}[1]{Chapter~\ref{#1}}
\providecommand{\sref}[1]{Section~\ref{#1}}
\providecommand{\fref}[1]{Figure~\ref{#1}}
\providecommand{\tref}[1]{Table~\ref{#1}}

\providecommand{\bignorm}[1]{\big\lVert#1\big\rVert}

\renewcommand{\vec}[1]{\ensuremath{\boldsymbol{#1}}}
\providecommand{\mat}[1]{\ensuremath{\boldsymbol{#1}}}

\providecommand{\calC}{\mathcal{C}}
\providecommand{\calD}{\mathcal{D}}
\providecommand{\calE}{\mathcal{E}}

\providecommand{\calL}{\mathcal{L}}

\providecommand{\calN}{\mathcal{N}}

\providecommand{\calP}{\mathcal{P}}

\providecommand{\calS}{\mathcal{S}}

\providecommand{\mE}{\mat{E}}
\providecommand{\mF}{\mat{F}}

\providecommand{\mI}{\mat{I}}

\providecommand{\mM}{\mat{M}}

\providecommand{\mW}{\mat{W}}

\providecommand{\vi}{\vec{i}}

\providecommand{\vn}{\vec{n}}

\providecommand{\vp}{\vec{p}}

\providecommand{\vy}{\vec{y}}

\usepackage{url}

\journal{Pattern Recognition Letters}

\begin{document}

\begin{frontmatter}

\title{Completely Weakly Supervised Class-Incremental Learning for Semantic Segmentation}

\author[1]{David Minkwan Kim}
\ead{kmk9846@hanyang.ac.kr}

\author[2]{Soeun Lee}
\ead{dlths67@cau.ac.kr}

\author[2]{Byeongkeun Kang\corref{cor1}}
\ead{byeongkeunkang@cau.ac.kr}
\cortext[cor1]{Corresponding author.}
\affiliation[1]{organization={Department of Computer Science, Hanyang University},
            addressline={222 Wangsimni-ro, Seongdong-gu}, 
            city={Seoul},
            postcode={04763}, 
            country={South Korea}}
\affiliation[2]{organization={School of Electrical and Electronics Engineering, Chung-Ang University},
            addressline={84 Heukseok-ro, Dongjak-gu}, 
            city={Seoul},
            postcode={06974}, 
            country={South Korea}}

\begin{abstract}
This work addresses the task of completely weakly supervised class-incremental learning for semantic segmentation to learn segmentation for both base and additional novel classes using only image-level labels. While class-incremental semantic segmentation (CISS) is crucial for handling diverse and newly emerging objects in the real world, traditional CISS methods require expensive pixel-level annotations for training. To overcome this limitation, partially weakly-supervised approaches have recently been proposed. However, to the best of our knowledge, this is the first work to introduce a completely weakly-supervised method for CISS. To achieve this, we propose to generate robust pseudo-labels by combining pseudo-labels from a localizer and a sequence of foundation models based on their uncertainty. Moreover, to mitigate catastrophic forgetting, we introduce an exemplar-guided data augmentation method that generates diverse images containing both previous and novel classes with guidance. Finally, we conduct experiments in three common experimental settings: 15-5 VOC, 10-10 VOC, and COCO-to-VOC, and in two scenarios: disjoint and overlap. The experimental results demonstrate that our completely weakly supervised method outperforms even partially weakly supervised methods in the 15-5 VOC and 10-10 VOC settings while achieving competitive accuracy in the COCO-to-VOC setting.
\end{abstract}


\end{frontmatter}




\section{Introduction}
\label{sec:intro}
Class-incremental semantic segmentation (CISS) has become a vital research topic in the computer vision and robotics communities~\cite{Shang2023Incrementer, Baek2022Decomposed, Cha2021SSUL}, as it enables learning to segment objects of novel classes in addition to previously learned categories using newly provided data. This capability is handy in various applications, including object manipulation, robot navigation, and industrial automation. For example, in object manipulation, CISS allows robots to learn to recognize and handle novel object categories not present in the initial training dataset. In robot navigation, CISS enables robots to incrementally learn about new types of obstacles and objects, allowing them to navigate dynamically changing environments safely and effectively.

Accordingly, many researchers have explored diverse methods to enhance accuracy in CISS~\cite{Shang2023Incrementer, Baek2022Decomposed, Cha2021SSUL}. However, these methods require dense pixel-level annotations in both the base and incremental learning steps, which are expensive and time-consuming. Specifically, for these methods, newly captured data must be annotated at the pixel level before using them for incremental learning. Meanwhile, to reduce annotation costs, researchers have studied weakly supervised~\cite{Zhao2024SFC} and semi-supervised~\cite{EAAI2025SemiSuper} learning methods for semantic segmentation.

To address this limitation, Cermelli \etal\cite{Cermelli2022Incremental} introduced a new task, weakly incremental learning for semantic segmentation (WILSS). In this task, a segmentation network is initially trained on base classes using pixel-level annotations. Then, it is incrementally updated to learn segmentation for novel classes using only image-level multi-labels while preserving its ability to segment previously learned categories. Because annotating image-level labels is significantly less time-consuming than obtaining dense labels~\cite{PRL2025Curriculum, PRL2023Weakly, PRL2022TRL, PRL2019Saliency, Kim2023Multiscale, Kang2024Improving}, the methods for WILSS are more practical and applicable in real-world scenarios than fully supervised CISS methods. 

\begin{figure}[!t] 
\centerline{\includegraphics[scale=0.9]{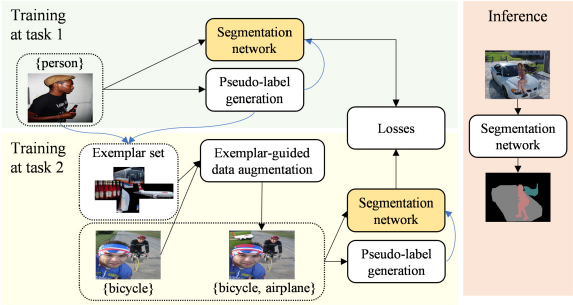}}
\caption{Overview of the proposed framework for completely weakly-supervised class-incremental learning for semantic segmentation.}
\label{fig:teaser}
\end{figure}

While WILSS methods~\cite{Cermelli2022Incremental, Roy2023RaSP, Yu2023Foundation, Kim2024Weakly} rely on densely annotated data to train a segmentation network on base classes, we argue that training the network using only weak supervision for both base and novel classes is crucial. It is to avoid dependence on specific densely annotated datasets and mitigate confusion caused by differences in class granularity and types across datasets. For instance, when a network is trained on a densely annotated dataset containing a broad class `furniture', incrementally training the network on novel classes such as `dining chair', `recliner', or `bar stool' may lead to inconsistencies. To address this limitation, we introduce a more challenging task, completely weakly supervised class-incremental semantic segmentation (CI-WSSS), as illustrated in~\fref{fig:teaser}.

To enable completely weakly supervised learning, we propose to generate dense pseudo-labels by combining pseudo-labels from a localizer and a sequence of foundation models. In addition to using the localizer as in~\cite{Cermelli2022Incremental, Roy2023RaSP, Yu2023Foundation, Kim2024Weakly}, we leverage a sequence of foundation models, including an open-set object detection model and a prompt-based image segmentation network. Then, based on the uncertainty of the pseudo-labels from the localizer, we combine the two pseudo-labels to produce more robust pseudo-labels.

To retain knowledge about previously learned classes, knowledge distillation losses are commonly employed~\cite{Cermelli2022Incremental, Roy2023RaSP, Yu2023Foundation, Kim2024Weakly}. Additionally, Yu \etal\cite{Yu2023Foundation} introduced a memory-based copy-paste augmentation method that stores a subset of previous training data (\ie exemplar set) and inserts a randomly sampled image from the set into newly provided data. While an exemplar set helps preserve knowledge about previous classes~\cite{Rebuffi2017iCaRL, fdcnet}, it typically contains a limited number of samples to constrain memory usage. Therefore, to increase the diversity of samples, we propose an exemplar-guided data augmentation method that blends an object image from the exemplar set with a current training image to generate a new image.

The contributions of this paper are summarized as follows: (1) We present, to the best of our knowledge, the first completely weakly supervised class-incremental semantic segmentation method that uses only image-level category labels to incrementally train a network for both base and novel classes. (2) To enable completely weakly supervised learning, we propose to combine the pseudo-labels from a localizer and a sequence of foundation models based on the uncertainty of the pseudo-labels to obtain more robust pseudo-labels. (3) To preserve knowledge of previously learned classes while learning new categories, we introduce an exemplar-guided data augmentation method that generates images containing both previous and novel classes with increased diversity. (4) We demonstrate the effectiveness of the proposed method in three common experimental settings: 15-5 VOC, 10-10 VOC, and COCO-to-VOC, and in two scenarios: disjoint and overlap.


\section{Related Work}
Cermelli \etal\cite{Cermelli2022Incremental} introduced a novel task, weakly-supervised incremental learning for semantic segmentation (WILSS). Given a semantic segmentation network trained on base classes using pixel-level labels, this task aims to incrementally train the network on novel classes using only image-level annotations. To achieve this, in each incremental step, Cermelli \etal\cite{Cermelli2022Incremental} trained a localizer using image-level class labels and dense predictions from the previous segmentation network. Subsequently, they extended the segmentation network and trained it using outputs from the localizer and the previous network.

Recently, Yu \etal\cite{Yu2023Foundation} presented a pseudo-label generation method that leverages foundation models for training a localizer. Specifically, they utilized a vision-language pre-trained model~\cite{Zhou2022Extract} and a self-supervised pre-trained model~\cite{zhou2022ibot}. Additionally, they introduced a memory-based copy-paste augmentation method, inspired by~\cite{Ghiasi2021Simple}. Roy \etal\cite{Roy2023RaSP} proposed to measure semantic closeness using text embeddings of class names. They then utilized the closeness to transfer the objectness prior of previously learned categories to new classes. Kim \etal\cite{Kim2024Weakly} proposed to utilize class hierarchy information in generating dense pseudo-labels. Liu \etal\cite{Liu2025Learning} introduced a framework that uses only web images to learn novel classes and preserve previously learned knowledge.

To the best of our knowledge, all previous WILSS methods~\cite{Cermelli2022Incremental, Yu2023Foundation, Roy2023RaSP, Kim2024Weakly} assume the availability of pixel-wise dense labels for base classes in the initial task. To overcome this limitation, we propose a class-incremental weakly-supervised semantic segmentation (CI-WSSS) method that incrementally learns to segment objects semantically using only image-level class labels for both base and novel classes.


\section{Method}
We first define the class-incremental WSSS (CI-WSSS) problem, which aims to achieve class-incremental semantic segmentation using only images and their image-level class labels. Then, we present a WSSS network that leverages online pseudo-labels from a localizer and offline pseudo-labels from a sequence of foundation models. Subsequently, we extend the WSSS framework to a CI-WSSS method. Additionally, we introduce an exemplar-guided data augmentation method to alleviate catastrophic forgetting.

\subsection{Problem Formulation} \label{sec:formulation}
At task $t=1$, a semantic segmentation network $f^1$ is trained using a dataset $\calD^1$ consisting of images and their image-level multi-labels where each image contains at least one of the base classes $\calC^1$. Later, at task $t=2$, the previous dataset $\calD^1$ is no longer available, and a new dataset $\calD^2$ is provided containing images and their image-level labels. Each image in $\calD^2$ includes at least one of the novel classes $\calC^2$. Given this, we train a network $f^2$ to segment objects of both $\calC^1$ and $\calC^2$ classes using $f^1$ and $\calD^2$. At a subsequent task $t > 2$, we train a network $f^t$ to segment objects of the accumulated classes $\calC^t_{\text{acc}} = \cup_{i=1}^{t} \calC^i$ using $f^{t-1}$ and $\calD^t$ containing $\calC^t$. Therefore, the network should incrementally learn segmentation of novel classes $\calC^t$ while retaining its knowledge about the previously learned classes $\calC^{t-1}_{\text{acc}}$.

\subsection{WSSS Network} \label{sec:method_wsss}
\begin{figure}[!t] 
\centering
\centerline{\includegraphics[scale=0.95]{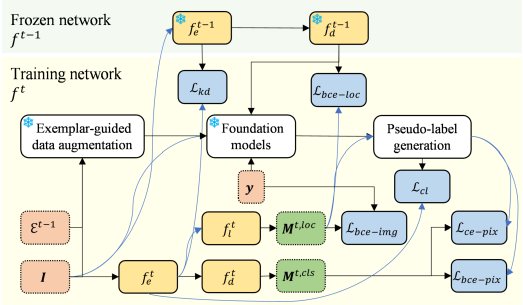}}
\caption{Proposed framework during training. Pink and green boxes denote inputs including the exemplar set ($\calE$) and outputs from the network, respectively. Yellow and blue boxes represent the modules in the network and the losses used for training, respectively.}
\label{fig:framework}
\end{figure}

\noindent \textbf{Network architecture}. 
At the first task $t=1$, our objective is to train a semantic segmentation network $f^1$ using a dataset $\calD^1$ that contains images and their image-level multi-labels to predict a class label for each pixel during inference. Accordingly, this initial task corresponds to WSSS using only image-level labels. A typical WSSS network consists of an encoder $f_e$, a decoder $f_d$, and a localizer $f_l$, as illustrated in~\fref{fig:framework}, where $f_l$ is employed only during training to generate dense pseudo-labels.

Given an image $\mI$, the encoder $f_e$ extracts a shared feature map $\mF$ and feeds it into both the decoder $f_d$ and the localizer $f_l$. Then, $f_d$ and $f_l$ produce a class score map $\mM^{cls} \in \mathbb{R}^{|\calC^1| \times H \times W}$ and a pseudo-label map $\mM^{\text{loc}} \in \mathbb{R}^{|\calC^1| \times H \times W}$, respectively, where $|\calC^1|$ denotes the number of classes in $\calC^1$; $H$ and $W$ are the height and width of $\mI$. We abbreviate the sequential process of $f_e$ and $f_d$ as $f_{e,d}$, and that of $f_e$ and $f_l$ as $f_{e,l}$ (\ie $f_{e,d}(\cdot) = f_d(f_e(\cdot))$ and $f_{e,l}(\cdot) = f_l(f_e(\cdot))$).

\vspace{1mm}
\noindent \textbf{Pseudo-label generation}.
We investigate the combined use of pseudo-labels $\mM^{fdt}$ generated from foundation models and pseudo-labels $\mM^{loc}$ from the localizer $f_l$. While foundation models have proven effective in tasks such as open-vocabulary image classification~\cite{Radford2021Learning} and detection~\cite{liu2023grounding}, their performance in open-set semantic segmentation is limited due to the lack of densely annotated data on the web.

Accordingly, to generate $\mM^{fdt}$, we employ a sequence of foundation models: an open-set detector and a prompt-based image segmentation model. Specifically, given an image $\mI$ and a text prompt constructed from its image-level class labels, we apply the open-set detector to generate bounding boxes for the objects corresponding to the labels. Subsequently, the predicted bounding boxes are utilized as prompts in the prompt-based image segmentation model to generate dense masks $\mM^{fdt} \in \mathbb{R}^{|\calC^1| \times H \times W}$.

\begin{figure}[!t] 
\centering
\centerline{\includegraphics[scale=0.19]{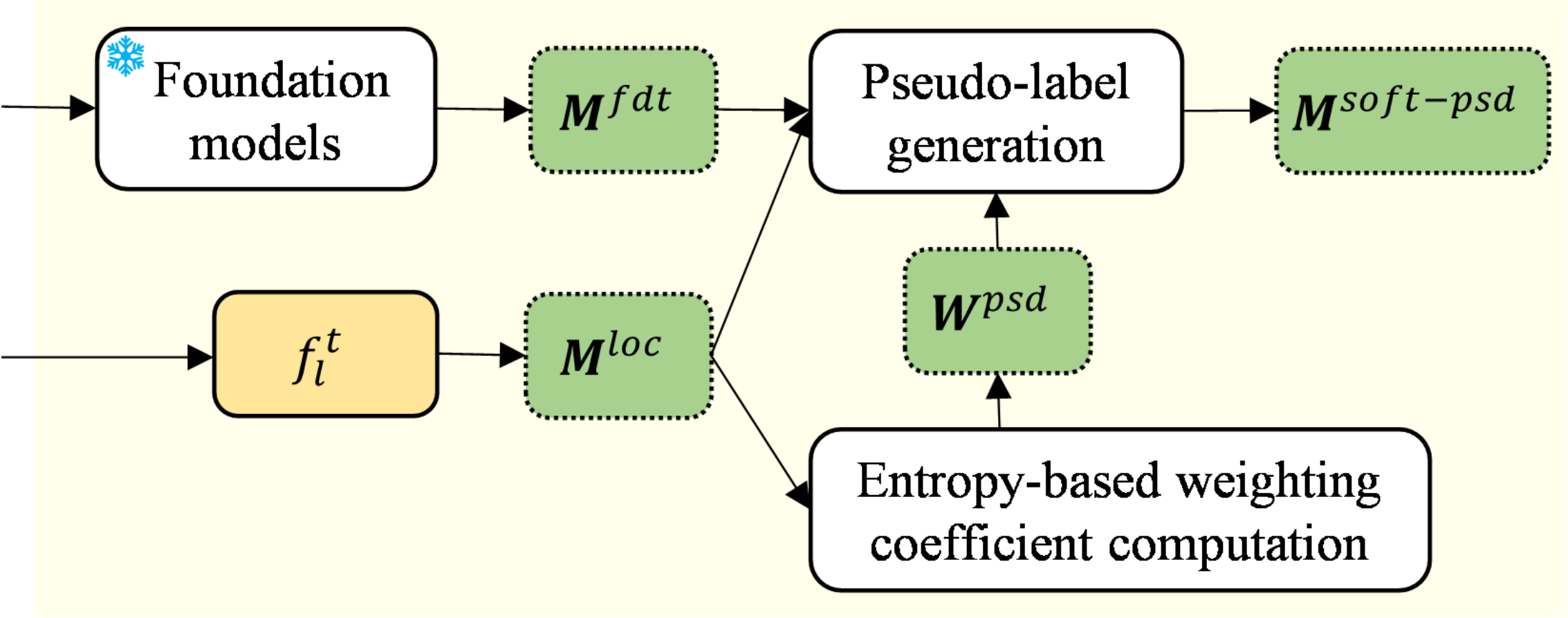}}
\caption{Soft pseudo-label generation. $\mM^{soft\text{-}psd}$, $\mW^{psd}$, $\mM^{fdt}$, and $\mM^{loc}$ denote the final soft pseudo-labels, entropy-based weighting coefficients, pseudo-labels from the foundation models, and those from the localizer $f_l$, respectively.}
\label{fig:framework_pseudo_label_gen}
\end{figure}

Lastly, based on the certainty of the localizer $f_l$ for each pixel, the final soft pseudo-labels $\mM^{soft\text{-}psd}$ are generated by a pixel-wise weighted summation of $\mM^{fdt}$ and $\mM^{loc}$, as presented in~\fref{fig:framework_pseudo_label_gen}. When the localizer has higher certainty for the class at ($h, w$), we increase reliance on $\mM^{loc}_{c,h,w}$, and vice versa. Certainty is quantified using the entropy $\mW^{psd}_{h,w}$ of the probability distribution for classes at ($h, w$). Specifically, the final pseudo-labels $\mM^{soft\text{-}psd}$ and the entropy $\mW^{psd}_{h,w}$ are computed as follows:
\begin{equation}
\begin{split}
& \mM^{soft\text{-}psd}_{c,h,w} := \mW^{psd}_{h,w} \mM^{fdt}_{c,h,w} + (1-\mW^{psd}_{h,w})  \sigma(\mM^{loc}_{c,h,w}), \\
& \mW^{psd}_{h,w} := \frac{-\sum_{c \in \calC^t} \bar{\mM}^{loc}_{c,h,w} \log(\bar{\mM}^{loc}_{c,h,w})}{\log (|\calC^t|)}
\end{split}
\end{equation}
where $\mW^{psd}_{h,w}$ is the entropy-based weighting coefficient at ($h, w$); $\sigma(\cdot)$ is a sigmoid function; $\bar{\mM}^{loc}_{c,h,w}$ is the output of the pixel-wise softmax function of $\mM^{loc}_{c,h,w}$. At the first task $t=1$, $\calC^t = \calC^1$. We employ Grounded DINO~\cite{liu2023grounding} for the open-set detector and SAM-HQ~\cite{Ke2023Segment} for the prompt-based image segmentation model, following~\cite{ren2024grounded}.

\vspace{1mm}
\noindent \textbf{Loss function}. 
The WSSS network is trained using the image-level multi-labels and the generated soft pseudo-labels $\mM^{soft\text{-}psd}$. We compute two pixel-wise cross-entropy losses by comparing the decoder output with $\mM^{soft\text{-}psd}$ and its modified version. Additionally, we employ a contrastive loss to pull together or push apart the features $\mF$ from the encoder based on pseudo-labels. Lastly, we calculate a loss by comparing the localizer output with the image-level labels to generate accurate pseudo-labels.

Given $\mM^{soft\text{-}psd}$, we train the encoder and decoder using a pixel-wise multi-class cross-entropy loss $\calL_{ce\text{-}pix}$ and a pixel-wise binary cross-entropy loss $\calL_{bce\text{-}pix}$. To compute $\calL_{ce\text{-}pix}$, we convert $\mM^{soft\text{-}psd}$ into hard pseudo-labels $\mM^{hard\text{-}psd}$ by finding the class with the maximum score at each location. Then, the loss $\calL_{ce\text{-}pix}$ and $\calL_{bce\text{-}pix}$ are computed as follows:
\begin{equation}
\begin{split}
\calL_{ce\text{-}pix} := - \frac{1}{HW} \sum_{c \in \calC} \sum_{h=1}^H \sum_{w=1}^W \mM^{hard\text{-}psd}_{c,h,w} \log(\mM^{cls}_{c,h,w}), \\
\calL_{bce\text{-}pix} := - \frac{1}{HW} \sum_{c \in \calC} \sum_{h=1}^H \sum_{w=1}^W [ \mM^{soft\text{-}psd}_{c,h,w} \log(\mM^{cls}_{c,h,w})  \\
+ (1 - \mM^{soft\text{-}psd}_{c,h,w}) \log(1 - \mM^{cls}_{c,h,w}) ]
\end{split}
\end{equation}
where $\mM^{cls}_{c,h,w}$ denotes the probability of the pixel at ($h,w$) belonging to class $c$, which is predicted by the decoder.

Additionally, we employ the dense contrastive loss $\calL_{cl}$ to train the encoder, following~\cite{Yu2023Foundation}. Specifically, we pull together the features of pixels with the same pseudo-labels while pushing apart those of pixels with different pseudo-labels. $\calL_{cl}$ is computed as follows:
\begin{equation}
\begin{split}
\calL_{cl} := -\frac{1}{|\calC|} & \sum_{c \in \calC} \frac{1}{|\calS_c|} \sum_{\vi \in \calS_c}  \frac{1}{|\calP_i|} \sum_{\vp \in \calP_i} \log \\
& \frac{\exp(\mF_{\vi}^\mathsf{T} \mF_{\vp} / \tau)}{\exp(\mF_{\vi}^\mathsf{T} \mF_{\vp} / \tau) + \sum_{\vn \in \calN_i} \exp(\mF_{\vi}^\mathsf{T} \mF_{\vn} / \tau)} 
\end{split}
\end{equation}
where $\calS_c$, $\calP_i$, and $\calN_i$ denote the sets containing the indexes of randomly sampled pixels with the pseudo-label $c$, the positive samples of pixel $i$, and the negative samples of $i$, respectively. The positive and negative samples are the pixels with the same and different hard pseudo-labels, respectively. $\tau$ is a temperature hyperparameter. 

Lastly, we use an image-level binary cross-entropy loss $\calL_{bce\text{-}img}$ in~\cite{Araslanov2020Single} to train the encoder and localizer. Specifically, we utilize the normalized Global Weighted Pooling (nGWP) $\vy^{\text{nGWP}}$ of the localizer output and the focal penalty term $\vy^{foc}$ to localize the entire object rather than only the most discriminative part. $\calL_{bce\text{-}img}$ is computed as follows:
\begin{equation}
\begin{split}
 \calL_{bce\text{-}img} & :=  - \sum_{c \in \calC} [\vy_c \log(\vp_c) + (1 - \vy_c) \log(1 - \vp_c)], \\
 \vp_c & :=  \sigma(\vy^{\text{nGWP}} + \vy^{foc}) \\
\end{split}
\end{equation}
where $\vy_c$ is the image-level label for class $c$; $\vp_c$ is the estimated probability of the image containing class $c$. For more details, we refer the readers to~\cite{Araslanov2020Single}.

The total loss $\calL_{wsss}$ for the WSSS network is computed as follows:
\begin{equation}
\begin{split}
\calL_{wsss} := \calL_{ce\text{-}pix} + \calL_{bce\text{-}img} + \alpha_1 \calL_{bce\text{-}pix} + \alpha_2 \calL_{cl} 
\end{split}
\label{eq:total_WSSS}
\end{equation}
where $\alpha_1$ and $\alpha_2$ are the hyperparameters to balance the losses. At the first task $t=1$, $\calC$ in the losses refers to $\calC^1$.

\subsection{Class-Incremental WSSS Network} \label{sec:method_ci_wsss}
Catastrophic forgetting is one of the main challenges in incremental learning~\cite{McCloskey1989Catastrophic, Goodfellow2013Catastrophic}. To mitigate this, we maintain a small subset of previous training data (\ie exemplar set) and propose an exemplar-guided data augmentation method that augments current training images with data from the exemplar set. Additionally, we employ a knowledge distillation loss to preserve the embedding space of the previous network in the current network. Lastly, we use the output of the previous network to guide the training of the localizer.

\vspace{1mm}
\noindent \textbf{Pseudo-label generation}.
The overall pseudo-label generation process in an incremental step remains consistent with the procedure described in~\sref{sec:method_wsss}. The main difference is that the text prompt for generating $\mM^{fdt}$ is constructed using both the ground-truth image-level labels and the predicted labels from the frozen network $f_{e,d}^{t-1}$. For an image $\mI \in \calD^t$ at the $t$-th task, its image-level labels are provided only for the novel classes $\calC^t$ while it may contain objects belonging to the previously learned classes $\calC^{t-1}_{\text{acc}}$. Therefore, we apply the frozen network $f_{e,d}^{t-1}$ to $\mI$ to obtain the predicted segmentation map and then incorporate all the predicted classes into the text prompt for the open-set detector. Because $f_{e,d}^{t-1}$ is frozen, we only need to apply it to all the images in $\calD^t$ once at the beginning of each incremental task. The generated pseudo-label maps $\mM^{soft\text{-}psd}$, $\mM^{fdt}$, and $\mM^{loc}$ have dimensions of $|\calC^{t}_{\text{acc}}| \times H \times W$.

\begin{figure}[!t] 
\centering
\centerline{\includegraphics[scale=0.19]{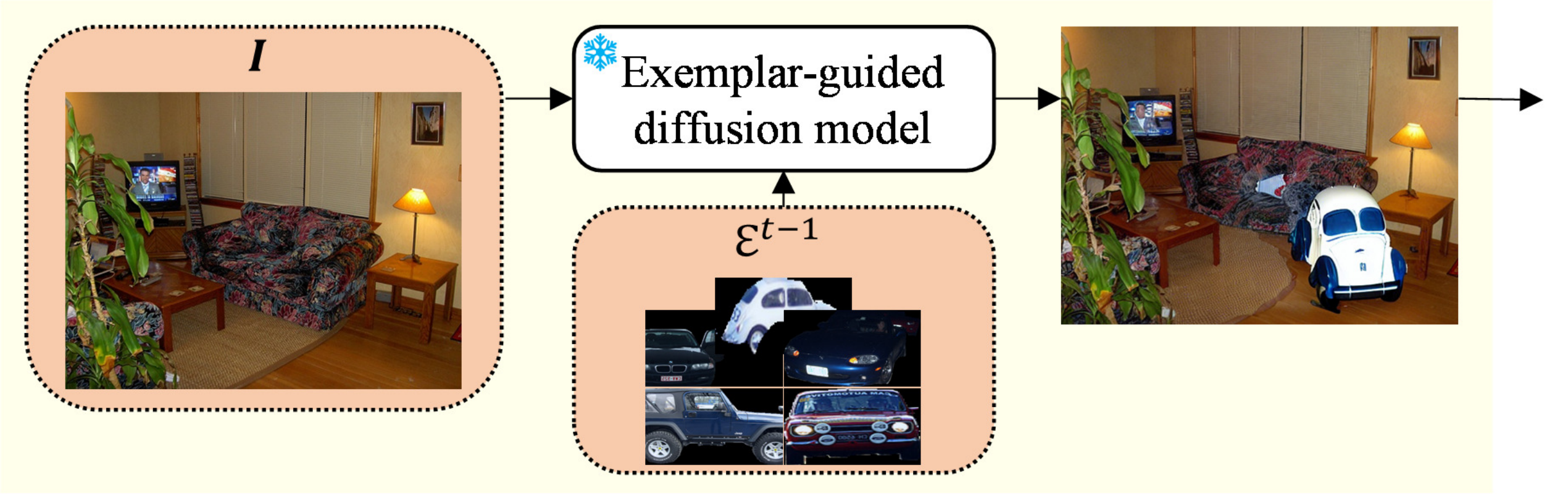}}
\caption{Exemplar-guided data augmentation. $\mI$ and $\calE^{t-1}$ represent a training image from the current task and the exemplar set from the previous task, respectively.}
\label{fig:framework_data_aug}
\end{figure}

\vspace{1mm}
\noindent \textbf{Exemplar-guided data augmentation}.
We investigate a method that maintains object images from previous tasks and utilizes them to augment current training images to retain the knowledge of previously learned classes $\calC^{t-1}_{\text{acc}}$ and enhance performance. We store object images of $\calC^{t-1}_{\text{acc}}$ in an exemplar set rather than entire images to efficiently maintain meaningful data. To crop object regions, we use pseudo-labels due to the lack of ground-truth dense labels. Then, we propose to leverage an exemplar-guided image editing diffusion model to augment current training images with object images from the exemplar set, as shown in~\fref{fig:framework_data_aug}. It is to increase the diversity of generated images for both the previously learned $\calC^{t-1}_{\text{acc}}$ and novel $\calC^{t}$ classes. Finally, We use the augmented images along with the current data for training.

\begin{figure}[!t]
\centering
\begin{minipage}{0.3\linewidth}
\centerline{\includegraphics[width=0.9\linewidth,height=0.08\textheight]{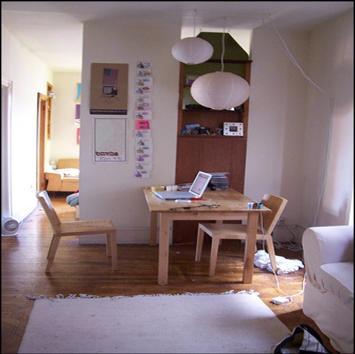}}
\end{minipage}
\begin{minipage}{0.3\linewidth}
\centerline{\includegraphics[width=0.9\linewidth,height=0.08\textheight]{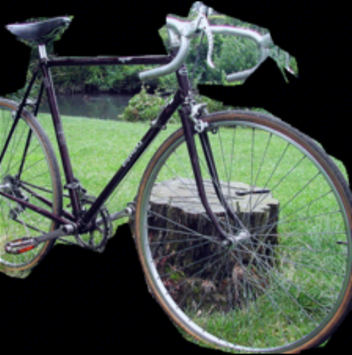}}
\end{minipage}
\begin{minipage}{0.3\linewidth}
\centerline{\includegraphics[width=0.9\linewidth,height=0.08\textheight]{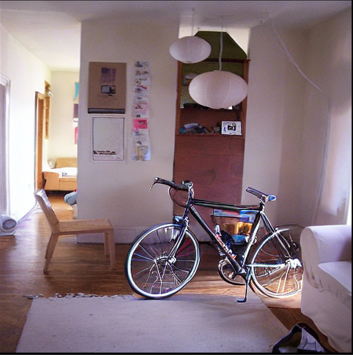}}
\end{minipage}
\\

\begin{minipage}{0.3\linewidth}
\centerline{\includegraphics[width=0.9\linewidth,height=0.08\textheight]{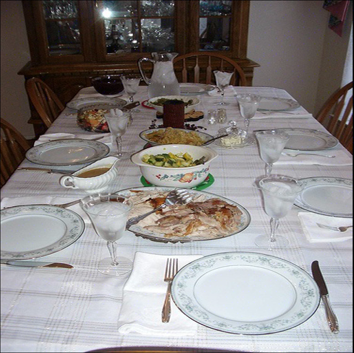}}
\end{minipage}
\begin{minipage}{0.3\linewidth}
\centerline{\includegraphics[width=0.9\linewidth,height=0.08\textheight]{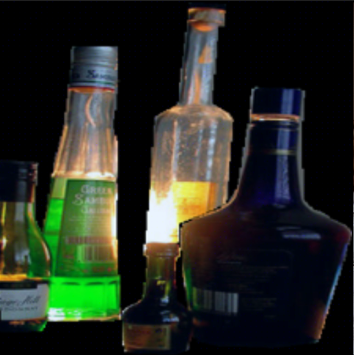}}
\end{minipage}
\begin{minipage}{0.3\linewidth}
\centerline{\includegraphics[width=0.9\linewidth,height=0.08\textheight]{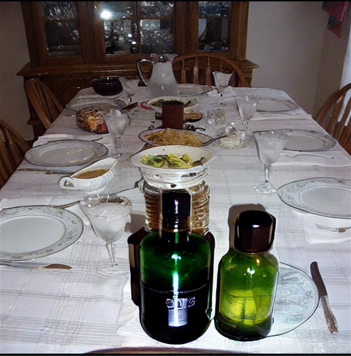}}
\end{minipage}
\caption{Results of exemplar-guided data augmentation. (Left) Current image; (Middle) Object image from an exemplar set; (Right) Result of augmentation.}
\label{fig:data_augmentation}
\end{figure}

Specifically, after the first task $t=1$, we construct an initial exemplar set $\calE^1$ by including randomly selected cropped object images of $\calC^1$ where object locations are obtained using pseudo-labels due to the unavailability of pixel-wise labels. In the subsequent task $t=2$, we randomly select an instance image $\mE$ from $\calE^1$ and a region in the current training image $\mI \in \calD^2$ for editing. Then, we apply the exemplar-guided image editing diffusion model to generate an image that includes the object from $\mE$ in the selected region of $\mI$, as shown in~\fref{fig:data_augmentation}. The inputs to the diffusion model are $\mI$, $\mE$, and a binary mask indicating the region. The synthesized images are processed by the pseudo-label generation method and used along with the data in $\calD^2$ during training. We use the method in~\cite{Yang2023Paint} for the exemplar-guided diffusion model.

\vspace{1mm}
\noindent \textbf{Loss function}. 
To alleviate catastrophic forgetting, we employ a knowledge distillation loss~\cite{Li2018LwF}, which encourages the feature maps $\mF^t$ from the current encoder $f_e^t$ to match those $\mF^{t-1}$ from the frozen previous encoder $f_e^{t-1}$ at the $t$-th task. The knowledge distillation loss $\calL_{kd}$ is computed as follows:
\begin{equation}
\begin{split}
\calL_{kd} := \bignorm{f_{e}^{t}(\mI) - f_{e}^{t-1}(\mI)}_2^2 = \bignorm{ \mF^t - \mF^{t-1} }_2^2.
\end{split}
\label{eq:loss_KD}
\end{equation}

Additionally, we use a binary cross-entropy loss to transfer knowledge from the previous network $f_{e,d}^{t-1}$ regarding the previously learned classes $\calC^{t-1}_{\text{acc}}$ to the current localizer $f_{e,l}^{t}$, facilitating the retention of the knowledge in the current network $f_{e,d}^{t}$. Accordingly, this loss minimizes the difference between the outputs of $f_{e,d}^{t-1}$ and the partial outputs of $f_{e,l}^{t}$ where the partial outputs correspond to $\calC^{t-1}_{\text{acc}}$. The loss $\calL_{bce\text{-}loc}$ is computed as follows:
\begin{equation}
\begin{split}
\calL_{bce\text{-}loc} := - \frac{1}{HW} & \sum_{c \in \calC^{t-1}_{\text{acc}}} \sum_{h=1}^H \sum_{w=1}^W [ \hat{\mM}^{t-1, cls}_{c,h,w} \log(\hat{\mM}^{t, loc}_{c,h,w}) \\
& + (1 - \hat{\mM}^{t-1, cls}_{c,h,w}) \log(1 - \hat{\mM}^{t, loc}_{c,h,w})]
\end{split}
\end{equation}
where $\hat{\mM}^{t-1, cls}$ and $\hat{\mM}^{t, loc}$ are the outputs of the sigmoid function applied to $\mM^{t-1, cls} \in \mathbb{R}^{|\calC^{t-1}_{\text{acc}}| \times H \times W }$ from the previous network and $\mM^{t, loc} \in \mathbb{R}^{|\calC^{t}_{\text{acc}}| \times H \times W }$ from the current localizer, respectively.

These two losses are used in addition to $\calL_{wsss}$ in~\eref{eq:total_WSSS} to train the class-incremental WSSS network. Therefore, the total loss $\calL_{ci\text{-}wsss}$ is computed as follows:
\begin{equation}
\calL_{ci\text{-}wsss} := \calL_{wsss} + \beta_1 \calL_{kd} + \beta_2 \calL_{bce\text{-}loc}
\label{eq:total_CI_WSSS}
\end{equation}
where $\beta_1$ and $\beta_2$ are the hyperparameters to balance the losses.


\section{Experiments and Results}
\subsection{Experimental Settings}
\noindent\textbf{Dataset and evaluation protocol.}
We conduct experiments on three common settings in WILSS: 15-5 VOC, 10-10 VOC, and COCO-to-VOC, following~\cite{Cermelli2022Incremental, Yu2023Foundation}. For the two VOC settings, we use the PASCAL VOC dataset~\cite{Everingham15Pascal}, containing 10,582 images for training and 1,449 images for validation. In the 15-5 VOC setting, we initially train the network on 15 base classes and then incrementally train it on the remaining 5 novel classes. For the 10-10 VOC setting, we split the 20 classes into 10 base classes and 10 novel classes.

Additionally, for each setting, we consider two scenarios: disjoint and overlap. In the disjoint scenario, the images in $\calD^t$ must contain at least one object belonging to the $\calC^t$ classes and may include objects from the $\calC^{t}_{\text{acc}}$ classes, where image-level labels are provided only for the $\calC^t$ classes. The images should not include objects from the novel classes $\cup_{i=t+1}^{T} \calC^i$ in future tasks. In the overlap scenario, the images in $\calD^t$ must contain at least one object from the $\calC^t$ classes with image-level labels and may include any additional objects without labels.  

In the COCO-to-VOC setting, the network is initially trained on the 60 COCO classes that do not overlap with the 20 VOC classes. We use the images in the COCO training set~\cite{coco}, which contain objects of only the 60 COCO classes. Subsequently, the network is trained on the 20 VOC classes using the entire VOC training set. We evaluate the network on both the COCO and VOC validation sets. Following~\cite{Cermelli2022Incremental, Yu2023Foundation}, we report evaluation results on the validation sets due to the unavailability of labels for the test sets.

Overall, we follow the experimental settings of WILSS~\cite{Cermelli2022Incremental, Yu2023Foundation}. However, we use only image-level labels for both the base and incremental steps, whereas WILSS~\cite{Cermelli2022Incremental, Yu2023Foundation} uses pixel-level labels for the initial task.

\vspace{1mm}
\noindent\textbf{Implementation details.}
We utilize DeepLabV3~\cite{Chen2017Rethinking} with the ResNet-101 backbone for the encoder $f_{e}$ and the decoder $f_{d}$, following~\cite{Cermelli2022Incremental, Yu2023Foundation}. The localizer $f_{l}$ consists of three convolution layers. In the VOC settings, at the initial task, we train the network for 40 epochs with a learning rate ($lr$) of 0.002, $\alpha_1=1$, and $\alpha_2=0.1$. At the incremental steps, we train the network for 40 epochs with $lr=0.00035$. We set $\alpha_1=1$, $\alpha_2=0.01$, $\beta_1=15$, and $\beta_2=1$ for the 15-5 VOC overlap scenario, $\alpha_1=2$ for the 15-5 VOC disjoint scenario, $\alpha_1=15$ for the 10-10 disjoint scenario, and $\alpha_1=10$, $\alpha_2=0.1$, and $\beta_1=1$ for the 10-10 overlap scenario, respectively. The omitted hyperparameters remain consistent with the 15-5 VOC overlap scenario.

In the COCO-to-VOC setting, we train the network for 40 epochs with $lr=0.01$, $\alpha_1=1$, and $\alpha_2=0.1$ at the initial task, and for 40 epochs with $lr=0.008$, $\alpha_1=1$, $\alpha_2=0.01$, $\beta_1=1$, and $\beta_2=0.5$ at the incremental step.

In the exemplar set, we store 50 object images per class, following~\cite{Yu2023Foundation}. To generate pseudo-labels $\mM^{fdt}$, we employ Grounding DINO~\cite{liu2023grounding} based on the Swin transformer~\cite{Liu2021Swin} and BERT~\cite{devlin2019bert} and SAM-HQ~\cite{Ke2023Segment} based on the MAE pre-trained vision transformer~\cite{He2016Deep}. The experiments were conducted on a server with four NVIDIA RTX A5000 GPUs, two Intel Xeon Silver 4215R CPUs, and 256 GB RAM. The code will be made available on GitHub upon publication to ensure reproducibility.

\begin{table}[!t]
\centering
\caption{Quantitative comparison in the 15-5 VOC setting.}
\label{tab:result_voc_15_5}
\scriptsize
\begin{tabular}{ >{\centering}m{0.075\textwidth}| >{\centering}m{0.04\textwidth}|>{\centering}m{0.0275\textwidth}|>{\centering}m{0.0325\textwidth}|>{\centering}m{0.025\textwidth}|>{\centering}m{0.0275\textwidth}|>{\centering}m{0.0325\textwidth}|>{\centering\arraybackslash}m{0.025\textwidth} } 
\hline
\multirow{2}{*}{Method} & \multirow{2}{*}{Label} & \multicolumn{3}{c|}{Disjoint} & \multicolumn{3}{c}{Overlap} \\
\cline{3-8}
  &  & 1-15 & 16-20 & All & 1-15 & 16-20 & All \\
\hline
Joint training & Pixel & 75.5 & 73.5 & 75.4 & 75.5 & 73.5 & 75.4 \\
\hline
CAM$^\dag$ &  & 69.3 & 26.1 & 59.4 & 69.9 & 25.6 & 59.7  \\
SEAM$^\dag$~\cite{Wang2020Self} & & 71.0 & 33.1 & 62.7 & 68.3 & 31.8 & 60.4  \\
SS$^\dag$~\cite{Araslanov2020Single} &  & 71.6 & 26.0 & 61.5 & 72.2 & 27.5 & 62.1  \\
EPS$^\dag$~\cite{Lee2023Saliency} & Pixel- & 72.4 & 38.5 & 65.2 & 69.4 & 34.5 & 62.1  \\
WILSON~\cite{Cermelli2022Incremental} & Image & 73.6 & 43.8 & 67.3 & 74.2 & 41.7 & 67.2  \\
WISH~\cite{Kim2024Weakly} &  & 74.4 & 45.1 & 68.2 & 75.3 & 45.4 & 68.9  \\
RaSP~\cite{Roy2023RaSP} & & - & - & - & 76.2 & 47.0 & 70.0  \\
FMWISS~\cite{Yu2023Foundation} & & \underline{75.9} & \underline{50.8} & \underline{70.7} & \underline{78.4} & \underline{54.5} & \underline{73.3} \\
\hline
Proposed & Image-Image & \textbf{77.3} & \textbf{52.2}& \textbf{72.1}& \textbf{78.8}& \textbf{56.9} & \textbf{74.2} \\
\hline
\end{tabular}
\end{table}

\begin{table}[!t]
\centering
\caption{Quantitative comparison in the 10-10 VOC setting.}
\label{tab:result_voc_10_10}
\scriptsize
\begin{tabular}{ >{\centering}m{0.075\textwidth}| >{\centering}m{0.04\textwidth}|>{\centering}m{0.0275\textwidth}|>{\centering}m{0.0325\textwidth}|>{\centering}m{0.025\textwidth}|>{\centering}m{0.0275\textwidth}|>{\centering}m{0.0325\textwidth}|>{\centering\arraybackslash}m{0.025\textwidth} } 
\hline
\multirow{2}{*}{Method} & \multirow{2}{*}{Label} & \multicolumn{3}{c|}{Disjoint} & \multicolumn{3}{c}{Overlap} \\
\cline{3-8}
  &   & 1-10 & 11-20 & All & 1-10 & 11-20 & All \\
\hline
Joint training & Pixel & 76.6 & 74.0 & 75.4 & 76.6 & 74.0 & 75.4 \\
\hline
CAM$^\dag$ &  & 65.4 & 41.3 & 54.5 & 70.8 & 44.2 & 58.5  \\
SEAM$^\dag$~\cite{Wang2020Self} &  & 65.1 & 53.5 & 60.6 & 67.5 & 55.4 & 62.7  \\
SS$^\dag$~\cite{Araslanov2020Single} &  & 60.7 & 25.7 & 45.0 & 69.6 & 32.8 & 52.5  \\
EPS$^\dag$~\cite{Lee2023Saliency} & Pixel- & 64.2 & 54.1 & 60.6 & 69.0 & 57.0 & 64.3  \\
WILSON~\cite{Cermelli2022Incremental} & Image & 64.5 & 54.3 & 60.8 & 70.4 & 57.1 & 65.0  \\
WISH~\cite{Kim2024Weakly} &  & 68.2 & 53.9 & 62.4 & 73.7 & 58.1 & 67.0  \\
RaSP~\cite{Roy2023RaSP} &  & - & - & - & 72.3 & 57.2 & 65.9  \\
FMWISS~\cite{Yu2023Foundation} &  & \underline{68.5} & \underline{58.2} & \underline{64.6} & \underline{73.8} & \underline{62.3} & \underline{69.1} \\
\hline
Proposed & Image-Image & \textbf{73.2} & \textbf{62.4}& \textbf{69.0}& \textbf{77.2}& \textbf{67.4}& \textbf{73.3} \\
\hline
\end{tabular}
\end{table}

\subsection{Results} \label{sec:result}
Tables~\ref{tab:result_voc_15_5},~\ref{tab:result_voc_10_10}, and~\ref{tab:result_coco_voc} present quantitative comparisons in mean Intersection over Union (mIoU) for the 15-5 VOC, 10-10 VOC, and COCO-to-VOC settings, respectively. The ``Joint training'' rows show results obtained using all training data throughout the entire process (\ie not incremental learning). Therefore, these results serve as upper bounds. In the second column, ``Pixel-Image'' denotes using pixel-level supervision for the initial task and image-level labels for the subsequent steps. ``Image-Image'' represents using image-level supervision for both initial and incremental tasks. The results marked with $^\dag$ are from~\cite{Cermelli2022Incremental}.

In the 15-5 and 10-10 VOC settings, our method outperforms the previous state-of-the-art (SOTA) method in both disjoint and overlap scenarios while using only image-level labels throughout the training process. In the COCO-to-VOC setting, although our method achieves higher accuracies on novel classes for both the COCO and VOC datasets than all previous works, it obtains lower accuracies on base classes than the previous SOTA method. 

In more detail, across all the settings, the proposed method outperforms the previous SOTA method for novel classes, where both methods are trained using only image-level labels. However, for base classes, the proposed method is trained with image-level labels, while the previous SOTA method uses pixel-level labels. Despite this, the proposed method still achieves higher accuracy on base classes in the VOC settings while slightly underperforming in the COCO-to-VOC setting. We believe that this is because images in the COCO dataset are more complex than those in the PASCAL VOC dataset, which leads to less accurate pseudo-labels for the COCO images. This, in turn, results in lower accuracy on base classes compared to the models trained with ground-truth dense labels.

\begin{table}[!t]
\centering
\caption{Quantitative comparison in the COCO-to-VOC setting.}
\label{tab:result_coco_voc}
\scriptsize
\begin{tabular}{ >{\centering}m{0.08\textwidth}| >{\centering}m{0.08\textwidth}|>{\centering}m{0.04\textwidth}|>{\centering}m{0.04\textwidth}|>{\centering}m{0.04\textwidth}|>{\centering\arraybackslash}m{0.04\textwidth} } 
\hline
\multirow{2}{*}{Method} & \multirow{2}{*}{Label} & \multicolumn{3}{c|}{COCO} & VOC \\
\cline{3-6}
  &  & 1-60 & 61-80 & All & 61-80 \\
\hline
CAM$^\dag$ &  & 30.7 & 20.3 & 28.1 & 39.1  \\
SEAM$^\dag$~\cite{Wang2020Self} & & 31.2 & 28.2 & 30.5 & 48.0  \\
SS$^\dag$~\cite{Araslanov2020Single} &  & 35.1 & 36.9 & 35.5 & 52.4  \\
EPS$^\dag$~\cite{Lee2023Saliency} & \multirow{2}{*}{Pixel-Image} & 34.9 & 38.4 & 35.8 & 55.3  \\
WILSON~\cite{Cermelli2022Incremental} &  & 39.8 & 41.0 & 40.6 & 55.7  \\
WISH~\cite{Kim2024Weakly} &  & \underline{40.2} & 43.0 & 41.5 & 58.7  \\
RaSP~\cite{Roy2023RaSP} &   & \textbf{41.1} & 40.7 & \textbf{41.6} & 54.4 \\
FMWISS~\cite{Yu2023Foundation} & & 39.9 & \underline{44.7} & \textbf{41.6} & \underline{63.6}  \\
\hline
Proposed & Image-Image & 38.2 & \textbf{45.0} & 40.4 & \textbf{64.0}  \\
\hline
\end{tabular}
\end{table}

\begin{figure}[!t]
  \centering 
\begin{minipage}{0.13\linewidth}
\centerline{\includegraphics[width=1.4\linewidth,height=0.06\textheight]{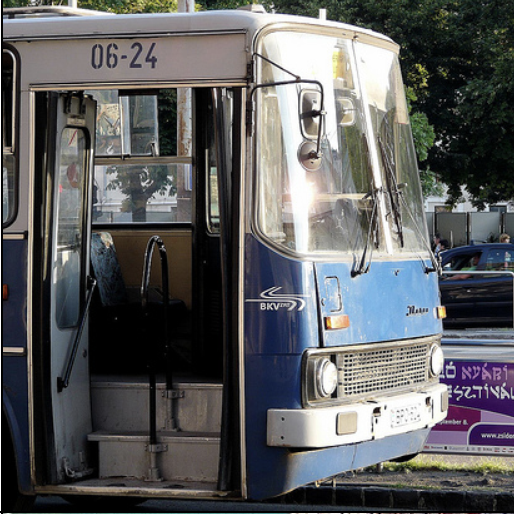}}
\end{minipage}
\hspace{0.3cm}
\begin{minipage}{0.13\linewidth}
\centerline{\includegraphics[width=1.4\linewidth,height=0.06\textheight]{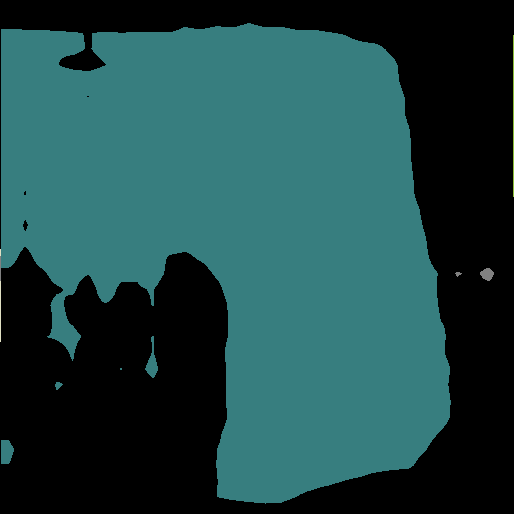}}
\end{minipage}
\hspace{0.3cm}
\begin{minipage}{0.13\linewidth}
\centerline{\includegraphics[width=1.4\linewidth,height=0.06\textheight]{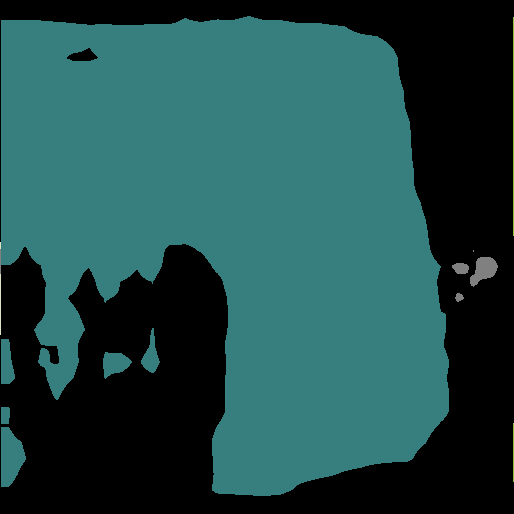}}
\end{minipage}
\hspace{0.3cm}
\begin{minipage}{0.13\linewidth}
\centerline{\includegraphics[width=1.4\linewidth,height=0.06\textheight]{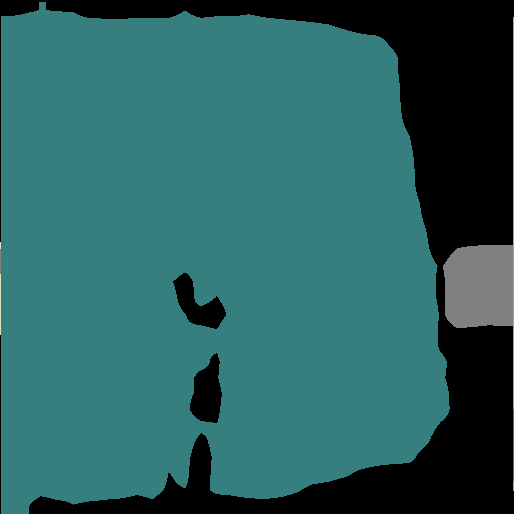}}
\end{minipage}
\hspace{0.3cm}
\begin{minipage}{0.13\linewidth}
\centerline{\includegraphics[width=1.4\linewidth,height=0.06\textheight]{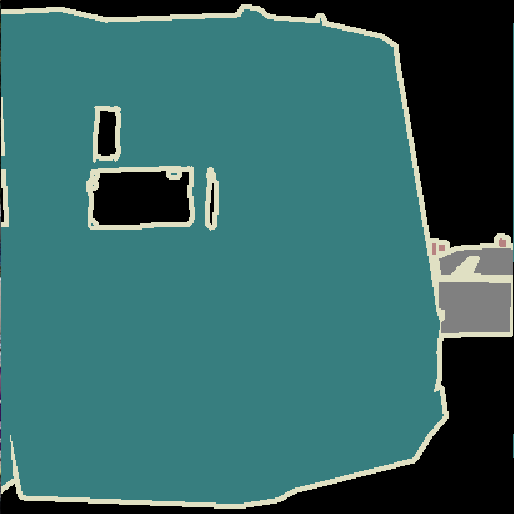}}
\end{minipage}
\\
\vspace{0.5mm}

\begin{minipage}{0.13\linewidth}
\centerline{\includegraphics[width=1.4\linewidth,height=0.06\textheight]{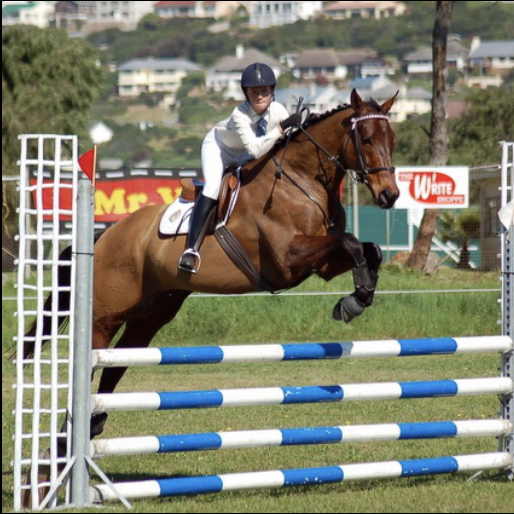}}
\end{minipage}
\hspace{0.3cm}
\begin{minipage}{0.13\linewidth}
\centerline{\includegraphics[width=1.4\linewidth,height=0.06\textheight]{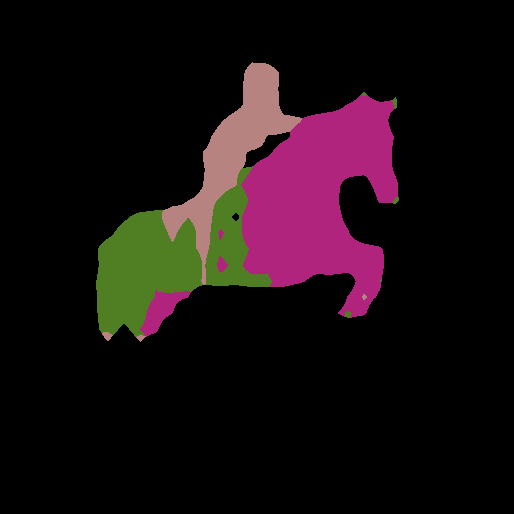}}
\end{minipage}
\hspace{0.3cm}
\begin{minipage}{0.13\linewidth}
\centerline{\includegraphics[width=1.4\linewidth,height=0.06\textheight]{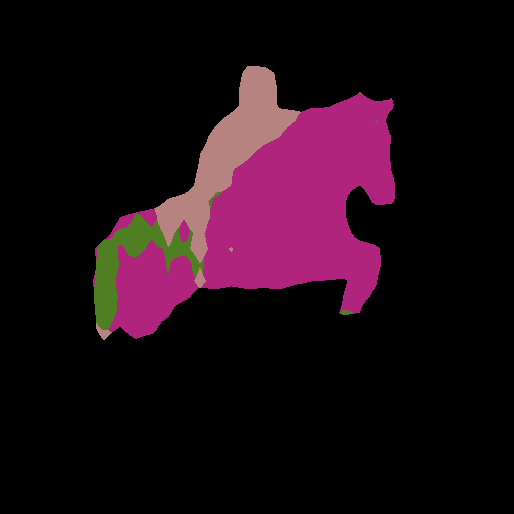}}
\end{minipage}
\hspace{0.3cm}
\begin{minipage}{0.13\linewidth}
\centerline{\includegraphics[width=1.4\linewidth,height=0.06\textheight]{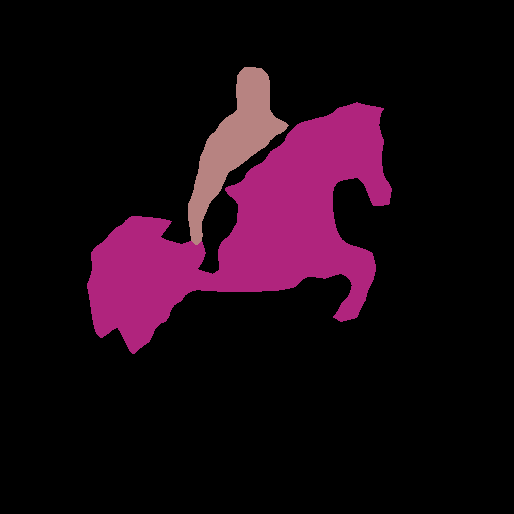}}
\end{minipage}
\hspace{0.3cm}
\begin{minipage}{0.13\linewidth}
\centerline{\includegraphics[width=1.4\linewidth,height=0.06\textheight]{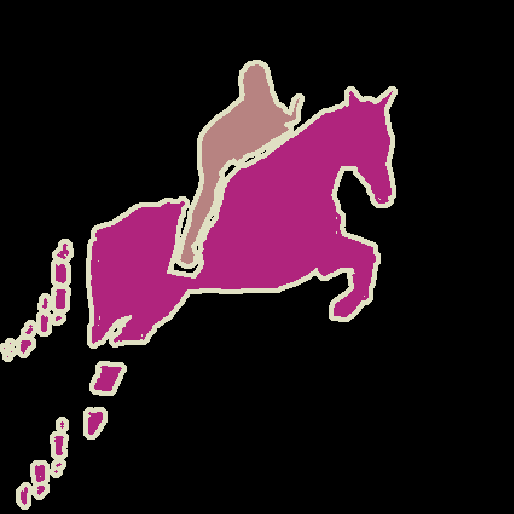}}
\end{minipage}
\\
\vspace{0.5mm}

\begin{minipage}{0.13\linewidth}
\centerline{\includegraphics[width=1.4\linewidth,height=0.06\textheight]{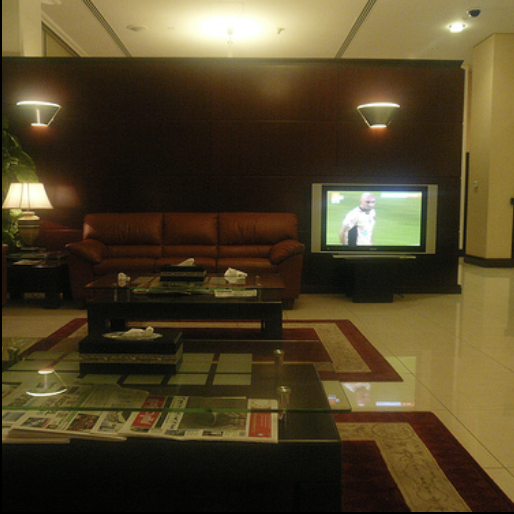}}
\end{minipage}
\hspace{0.3cm}
\begin{minipage}{0.13\linewidth}
\centerline{\includegraphics[width=1.4\linewidth,height=0.06\textheight]{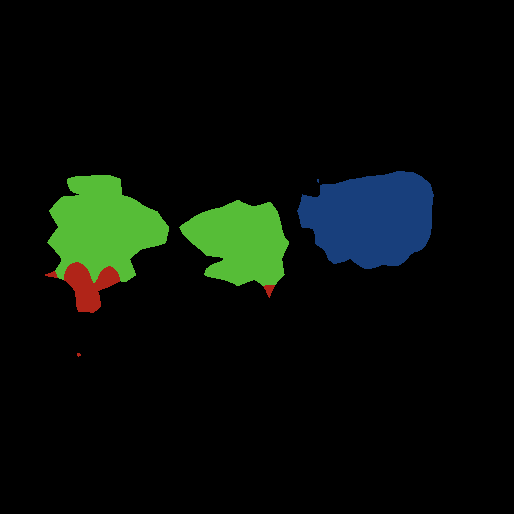}}
\end{minipage}
\hspace{0.3cm}
\begin{minipage}{0.13\linewidth}
\centerline{\includegraphics[width=1.4\linewidth,height=0.06\textheight]{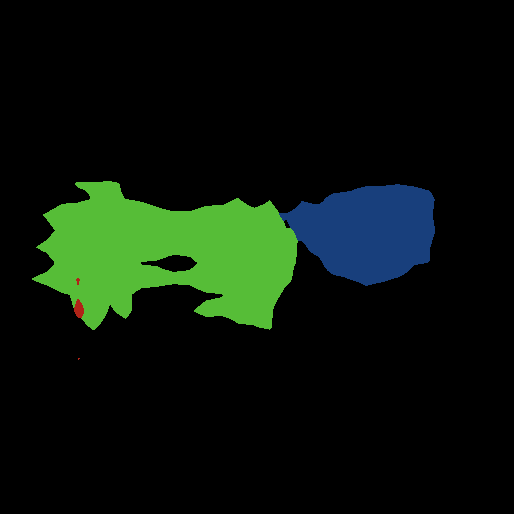}}
\end{minipage}
\hspace{0.3cm}
\begin{minipage}{0.13\linewidth}
\centerline{\includegraphics[width=1.4\linewidth,height=0.06\textheight]{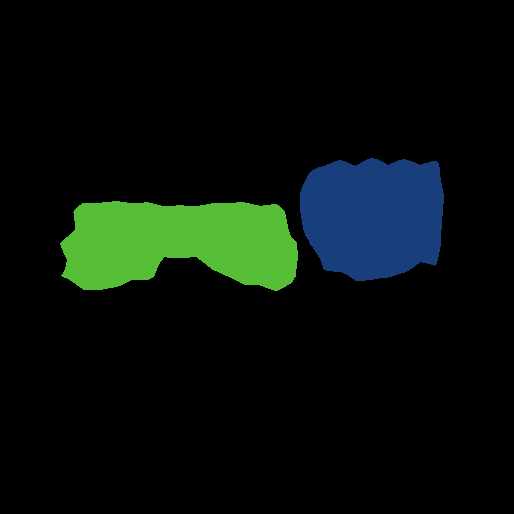}}
\end{minipage}
\hspace{0.3cm}
\begin{minipage}{0.13\linewidth}
\centerline{\includegraphics[width=1.4\linewidth,height=0.06\textheight]{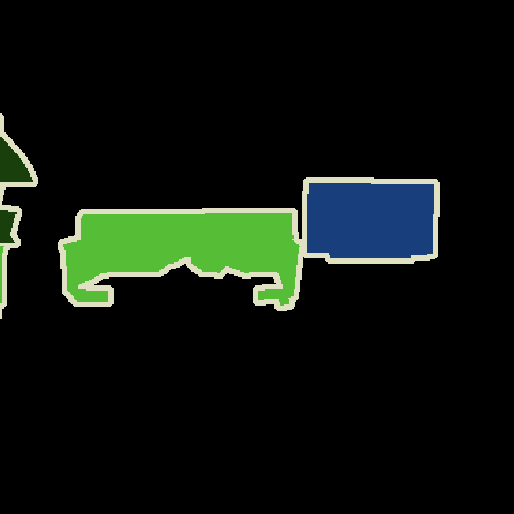}}
\end{minipage}
\\
\vspace{0.5mm}

\begin{minipage}{0.13\linewidth}
\centerline{\footnotesize (a)}
\end{minipage}
\hspace{0.3cm}
\begin{minipage}{0.13\linewidth}
\centerline{\footnotesize (b)}
\end{minipage}
\hspace{0.3cm}
\begin{minipage}{0.13\linewidth}
\centerline{\footnotesize (c)}
\end{minipage}
\hspace{0.3cm}
\begin{minipage}{0.13\linewidth}
\centerline{\footnotesize (d)}
\end{minipage}
\hspace{0.3cm}
\begin{minipage}{0.13\linewidth}
\centerline{\footnotesize (e)}
\end{minipage}
\caption{Qualitative comparison in the 10-10 VOC setting. (a) Image; (b) WILSON~\cite{Cermelli2022Incremental}; (c) FMWISS~\cite{Yu2023Foundation}; (d) Ours; (e) Ground-truth.}
\label{fig:result_10_10}
\end{figure}

\begin{figure}[!t]
  \centering
\begin{minipage}{0.13\linewidth}
\centerline{\includegraphics[width=1.4\linewidth,height=0.06\textheight]{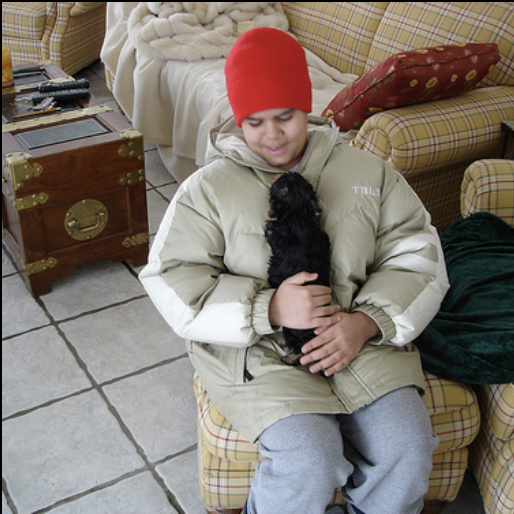}}
\end{minipage}
\hspace{0.3cm}
\begin{minipage}{0.13\linewidth}
\centerline{\includegraphics[width=1.4\linewidth,height=0.06\textheight]{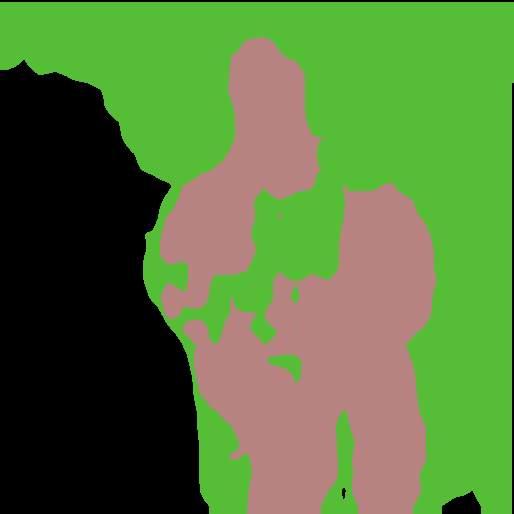}}
\end{minipage}
\hspace{0.3cm}
\begin{minipage}{0.13\linewidth}
\centerline{\includegraphics[width=1.4\linewidth,height=0.06\textheight]{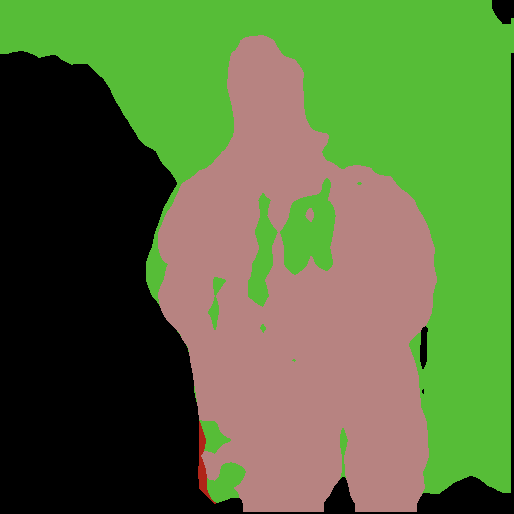}}
\end{minipage}
\hspace{0.3cm}
\begin{minipage}{0.13\linewidth}
\centerline{\includegraphics[width=1.4\linewidth,height=0.06\textheight]{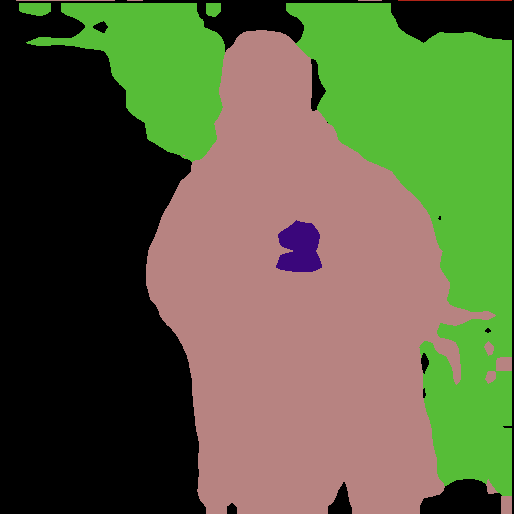}}
\end{minipage}
\hspace{0.3cm}
\begin{minipage}{0.13\linewidth}
\centerline{\includegraphics[width=1.4\linewidth,height=0.06\textheight]{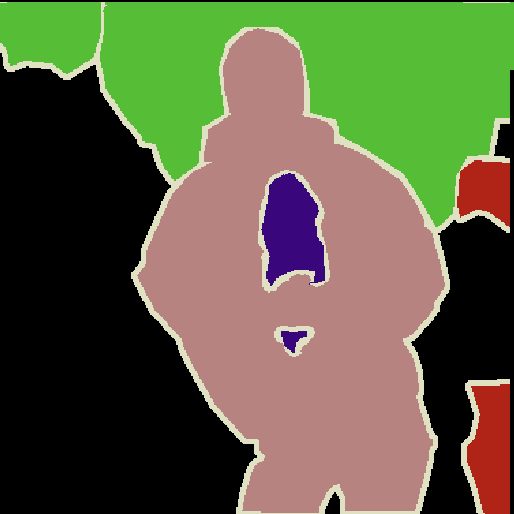}}
\end{minipage}
\\
\vspace{0.5mm}

\begin{minipage}{0.13\linewidth}
\centerline{\includegraphics[width=1.4\linewidth,height=0.06\textheight]{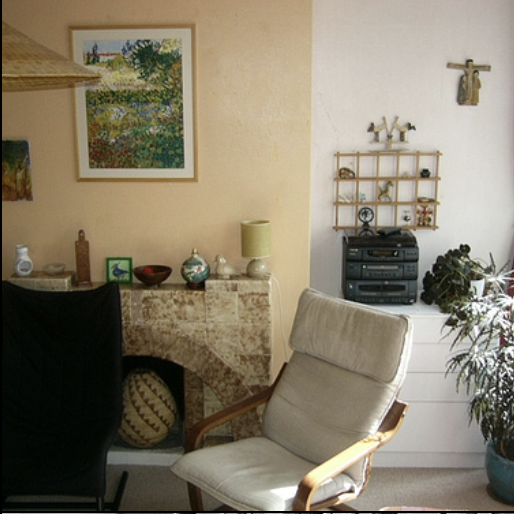}}
\end{minipage}
\hspace{0.3cm}
\begin{minipage}{0.13\linewidth}
\centerline{\includegraphics[width=1.4\linewidth,height=0.06\textheight]{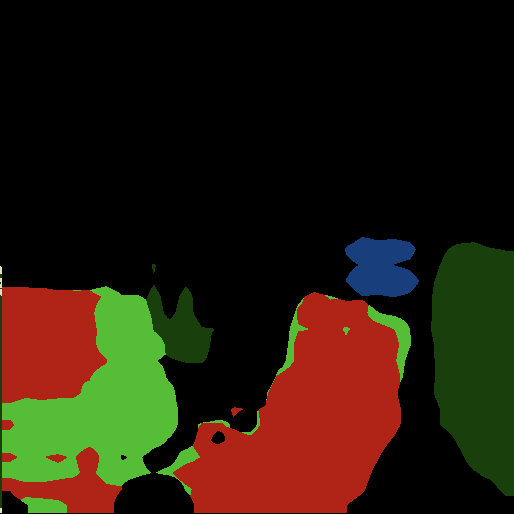}}
\end{minipage}
\hspace{0.3cm}
\begin{minipage}{0.13\linewidth}
\centerline{\includegraphics[width=1.4\linewidth,height=0.06\textheight]{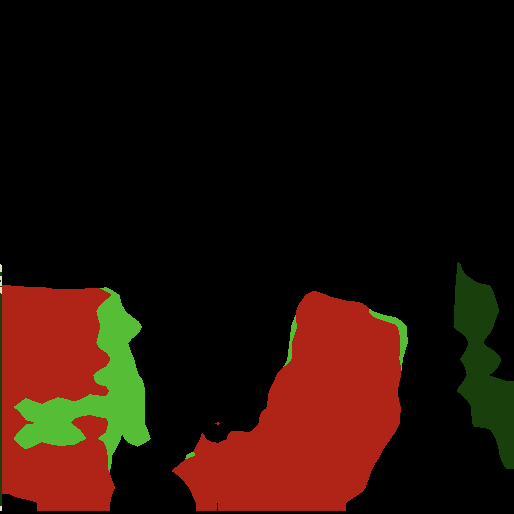}}
\end{minipage}
\hspace{0.3cm}
\begin{minipage}{0.13\linewidth}
\centerline{\includegraphics[width=1.4\linewidth,height=0.06\textheight]{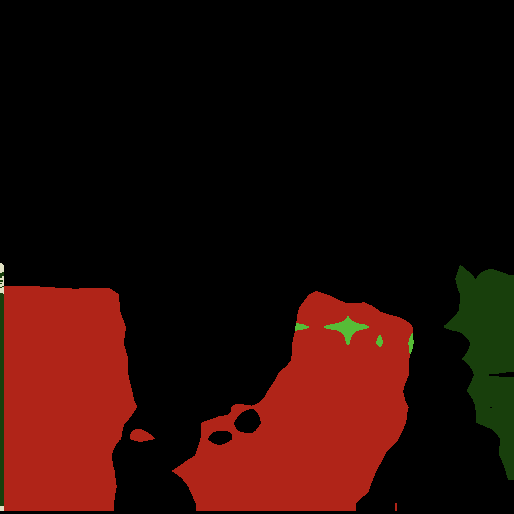}}
\end{minipage}
\hspace{0.3cm}
\begin{minipage}{0.13\linewidth}
\centerline{\includegraphics[width=1.4\linewidth,height=0.06\textheight]{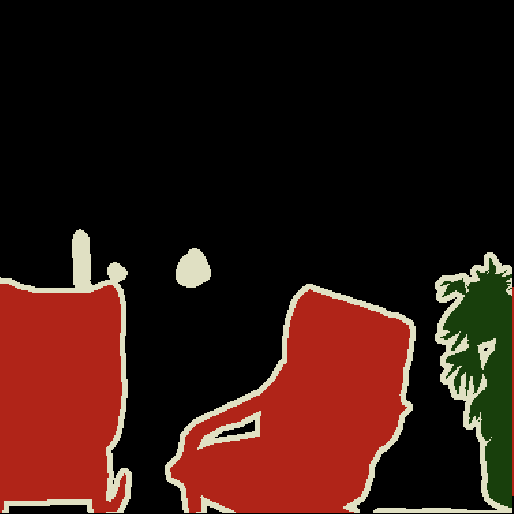}}
\end{minipage}
\\
\vspace{0.5mm}

\begin{minipage}{0.13\linewidth}
\centerline{\includegraphics[width=1.4\linewidth,height=0.06\textheight]{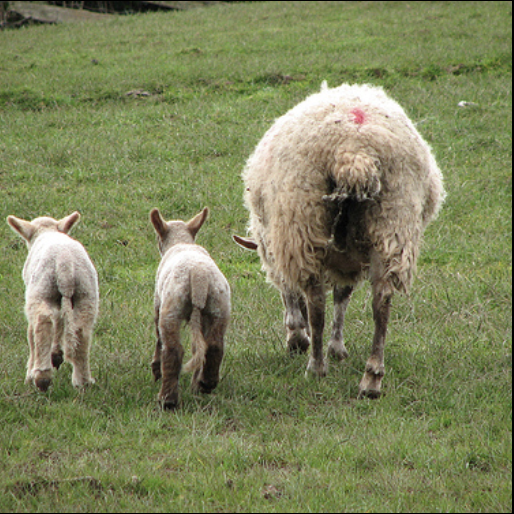}}
\end{minipage}
\hspace{0.3cm}
\begin{minipage}{0.13\linewidth}
\centerline{\includegraphics[width=1.4\linewidth,height=0.06\textheight]{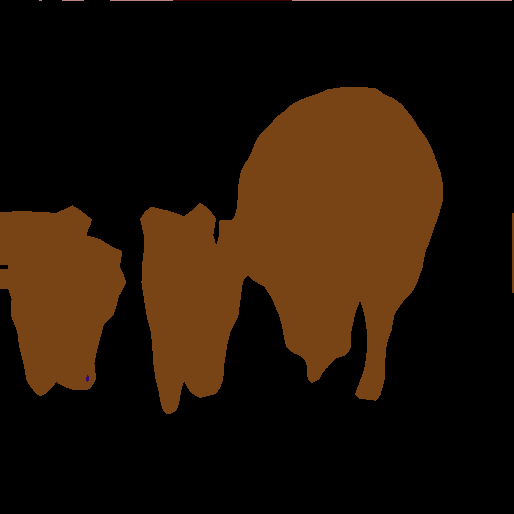}}
\end{minipage}
\hspace{0.3cm}
\begin{minipage}{0.13\linewidth}
\centerline{\includegraphics[width=1.4\linewidth,height=0.06\textheight]{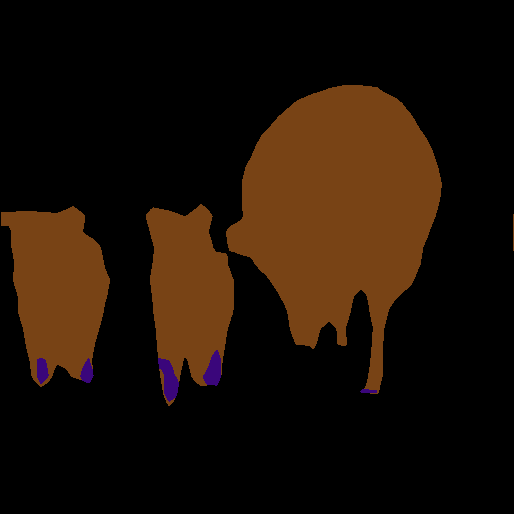}}
\end{minipage}
\hspace{0.3cm}
\begin{minipage}{0.13\linewidth}
\centerline{\includegraphics[width=1.4\linewidth,height=0.06\textheight]{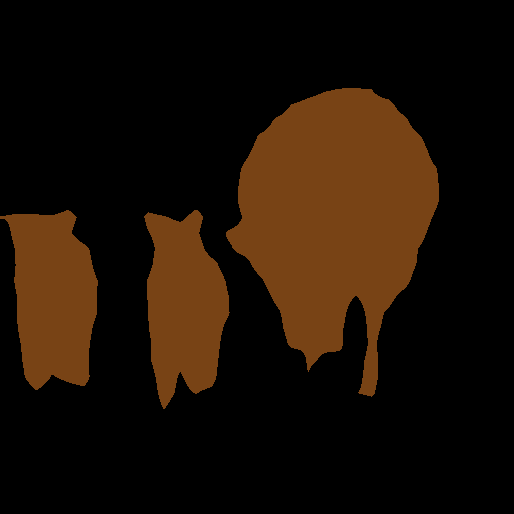}}
\end{minipage}
\hspace{0.3cm}
\begin{minipage}{0.13\linewidth}
\centerline{\includegraphics[width=1.4\linewidth,height=0.06\textheight]{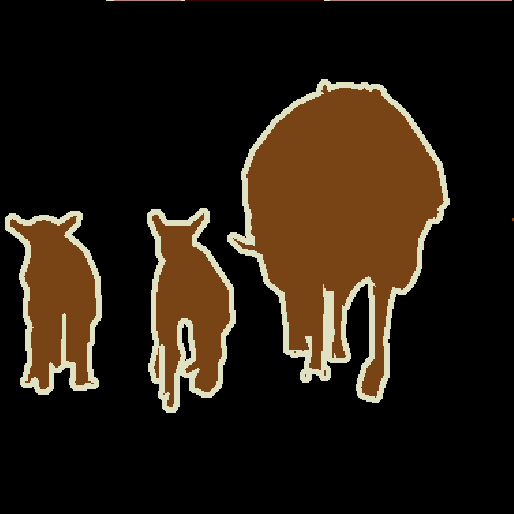}}
\end{minipage}
\\
\vspace{0.5mm}

\begin{minipage}{0.13\linewidth}
\centerline{\footnotesize (a)}
\end{minipage}
\hspace{0.3cm}
\begin{minipage}{0.13\linewidth}
\centerline{\footnotesize (b)}
\end{minipage}
\hspace{0.3cm}
\begin{minipage}{0.13\linewidth}
\centerline{\footnotesize (c)}
\end{minipage}
\hspace{0.3cm}
\begin{minipage}{0.13\linewidth}
\centerline{\footnotesize (d)}
\end{minipage}
\hspace{0.3cm}
\begin{minipage}{0.13\linewidth}
\centerline{\footnotesize (e)}
\end{minipage}
\caption{Qualitative comparison in the 15-5 VOC setting. (a) Image; (b) WILSON~\cite{Cermelli2022Incremental}; (c) FMWISS~\cite{Yu2023Foundation}; (d) Ours; (e) Ground-truth.}
\label{fig:result_15_5}
\end{figure}

Figures~\ref{fig:result_10_10} and~\ref{fig:result_15_5} show qualitative comparisons for the 10-10 VOC and 15-5 VOC settings, respectively. The results demonstrate that the proposed method predicts more accurate semantic segmentation maps than previous works~\cite{Cermelli2022Incremental, Yu2023Foundation} for both base and novel classes. In~\fref{fig:result_10_10}, \{\textit{bus, car}\} and \{\textit{horse, person, sofa, tv-monitor}\} are base classes and novel classes, respectively. In~\fref{fig:result_15_5}, \{\textit{person, chair, dog}\} and \{\textit{sheep, sofa, plant}\} belong to base classes and novel classes, respectively.

\subsection{Analysis} \label{sec:analysis}
We present an ablation study on pseudo-label generation using foundation models and exemplar-guided data augmentation in~\tref{tab:ablation_voc}. We first train a WSSS network on base classes as described in~\sref{sec:method_wsss}, including pseudo-label generation using foundation models. Then, we incrementally train the network on novel classes using the method presented in~\sref{sec:method_ci_wsss}, but excluding some components. The first row shows the results of the baseline method, which does not use pseudo-labels $\mM^{fdt}$ from foundation models and the exemplar set $\calE$. The second row presents the results of incorporating $\mM^{fdt}$ into the baseline method. These results demonstrate that dense pseudo-labels $\mM^{fdt}$ for novel classes significantly enhance accuracy, especially for the novel classes. The last row shows the results of additionally employing exemplar-guided data augmentation. The results indicate that the augmentation method helps the model retain knowledge about previously learned classes.

\begin{table}[!t]
\centering
\caption{Ablation study on pseudo-label ($\mM^{fdt}$) and exemplar-guided data augmentation in the 15-5 VOC setting.}
\label{tab:ablation_voc}
\scriptsize
\begin{minipage}{1\linewidth}
\centering
\begin{tabular}{ >{\centering}m{0.1\textwidth}| >{\centering}m{0.16\textwidth}| >{\centering}m{0.0525\textwidth}| >{\centering}m{0.0675\textwidth}| >{\centering}m{0.045\textwidth}| >{\centering}m{0.0525\textwidth}| >{\centering}m{0.0675\textwidth}| >{\centering\arraybackslash}m{0.045\textwidth} } 
\hline
\multicolumn{2}{c|}{Method} & \multicolumn{3}{c|}{Disjoint} & \multicolumn{3}{c}{Overlap} \\
\hline
Pseudo-label ($\mM^{fdt}$) & Exemplar-guided augmentation & 1-15 & 16-20 & All & 1-15 & 16-20 & All \\
\hline
- & - & 75.8 & 47.2 & 69.8 & 76.8 & 50.7 & 71.3 \\
\checkmark & - & 76.4 & \textbf{54.2} & 71.9 & 77.5 & 55.4 & 73.0 \\
\checkmark & \checkmark & \textbf{77.3} & 52.2 & \textbf{72.1}& \textbf{78.8} & \textbf{56.9} & \textbf{74.2} \\
\hline
\end{tabular}
\end{minipage}
\end{table}

\begin{table}[!t]
\centering
\caption{Comparison of pseudo-label generation methods in the 10-10 VOC setting.}
\label{tab:analysis_pseudo_label}
\scriptsize
\begin{minipage}{1\linewidth}
\centering
\begin{tabular}{ >{\centering}m{0.23\textwidth}| *5{>{\centering}m{0.07\textwidth}|} >{\centering\arraybackslash}m{0.07\textwidth} } 
\hline
\multirow{2}{*}{Method for $\mM^{fdt}$} & \multicolumn{3}{c|}{Disjoint} & \multicolumn{3}{c}{Overlap} \\
\cline{2-7}
  &1-10 & 11-20 & All & 1-10 & 11-20 & All \\
\hline
FMWISS~\cite{Yu2023Foundation} & 54.9 & 55.4 & 56.8 & 55.1 & 52.1 & 55.2 \\
Proposed & \textbf{73.2} & \textbf{62.4} & \textbf{69.0} & \textbf{77.2} & \textbf{67.4} & \textbf{73.3} \\
\hline
\end{tabular}
\end{minipage}
\end{table}

\begin{table}[!t]
\centering
\caption{Comparison of augmentation methods in the 10-10 VOC setting.}
\label{tab:analysis_augmentation}
\scriptsize
\begin{minipage}{1\linewidth}
\centering
\begin{tabular}{ >{\centering}m{0.23\textwidth}| *5{>{\centering}m{0.07\textwidth}|} >{\centering\arraybackslash}m{0.07\textwidth} } 
\hline
\multirow{2}{*}{Method} & \multicolumn{3}{c|}{Disjoint} & \multicolumn{3}{c}{Overlap} \\
\cline{2-7}
  & 1-10 & 11-20 & All & 1-10 & 11-20 & All \\
\hline
Copy-paste in~\cite{Yu2023Foundation} & 72.9 & 60.8 & 68.1 & 76.8 & 66.5 & 72.6 \\
Proposed & \textbf{73.2} & \textbf{62.4} & \textbf{69.0}& \textbf{77.2} & \textbf{67.4} & \textbf{73.3} \\
\hline
\end{tabular}
\end{minipage}
\end{table}

In~\tref{tab:analysis_pseudo_label}, we compare the results of using the proposed pseudo-label generation method with those of the method in~\cite{Yu2023Foundation}. In the first row, we replace $\mM^{fdt}$ in Sections~\ref{sec:method_wsss} and~\ref{sec:method_ci_wsss} with the pseudo-labels used in~\cite{Yu2023Foundation} while maintaining all other components of our method. The results confirm that the proposed pseudo-label generation method is more effective than the method in~\cite{Yu2023Foundation}.

\tref{tab:analysis_augmentation} shows quantitative comparisons of the results using the proposed exemplar-guided data augmentation method and those using the copy-paste method in~\cite{Yu2023Foundation}. For fair comparisons, we constructed the exemplar set $\calE$ using the same images in each scenario. For the results of the first row, we replaced our augmentation method with the copy-paste method from~\cite{Yu2023Foundation} while preserving others. The results demonstrate that the proposed method outperforms the models trained using the copy-paste method in~\cite{Yu2023Foundation}. We believe this is because our model generates images with greater diversity for both base and novel classes.

Tables~\ref{tab:analysis_pseudo_label} and~\ref{tab:analysis_augmentation} show the advantages of the proposed pseudo-label generation and augmentation methods compared to those in the state-of-the-art method~\cite{Yu2023Foundation} while \tref{tab:ablation_voc} presents comparisons against our baseline model. First, the experimental results demonstrate that the proposed pseudo-label generation method outperforms the one in~\cite{Yu2023Foundation}. Our approach combines pseudo-labels from the localizer and the sequence of foundation models based on the uncertainty of the pseudo-labels from the localizer. Accordingly, in regions where the localizer is uncertain, the pseudo-labels from the sequence of foundation models contribute more to generating the final pseudo-labels. We believe that this uncertainty-based fusion strategy effectively leverages both pseudo-labels and generates more robust pseudo-labels.

Moreover, the results also show that the proposed augmentation method is superior to the one in~\cite{Yu2023Foundation}. As presented in~\fref{fig:data_augmentation}, our exemplar-guided data augmentation method generates more realistic and diverse images containing both previously learned and novel objects than the copy-paste augmentation method in~\cite{Yu2023Foundation}. These more realistic and diverse images contribute to improving the performance of the proposed model.

\subsection{Discussion} \label{sec:discussion}
While the proposed method achieves promising results as presented in Tables~\ref{tab:result_voc_15_5},~\ref{tab:result_voc_10_10}, and~\ref{tab:result_coco_voc}, it might encounter certain challenges during practical deployment. First, although the proposed approach significantly reduces annotation costs by relying only on image-level labels and leveraging foundation models, the generated pixel-level pseudo-labels might be inaccurate in rare or specialized environments such as space exploration. Additionally, the quality of the exemplar set depends on the accuracy of the pixel-level pseudo-labels for previously learned classes. Since object regions are cropped based on these pseudo-labels to construct the exemplar set, any inaccuracies in the pseudo-labels can result in exemplar sets with poor-quality dense labels. These challenges stem from the inherent difficulties of weakly supervised learning, particularly due to the absence of pixel-level annotations.

Moreover, since the proposed method leverages foundation models and an exemplar-guided diffusion model, it requires additional computation and memory during training. Depending on the application, these additional requirements may be costly or infeasible. Lastly, to learn a novel object category, we assume that a sufficient number of images containing the novel objects are available. Specifically, we do not address scenarios where only a few images of the novel class are present.

\section{Conclusion}
We introduced the task of completely weakly-supervised class-incremental semantic segmentation, which aims to learn semantic segmentation on base classes and incrementally train on novel classes, all using only image-level class labels while maintaining the ability to segment previously learned classes. We then presented a robust pseudo-label generation method that combines pseudo-labels from a localizer and foundation models based on their uncertainty. We also introduced an exemplar-guided data augmentation method to generate diverse images containing objects of both previous and novel classes. Experimental results demonstrate that our completely weakly-supervised method outperforms or achieves competitive accuracy compared to partially weakly-supervised approaches.

Regarding limitations, the proposed method assumes that when the pseudo-labels from the localizer are unreliable, the pseudo-labels from the foundation models are more accurate. While this assumption usually holds for the images in our experiments, it may not apply to images from particular domains.

\section*{Acknowledgments}
This research was supported in part by the National Research Foundation of Korea (NRF) grant funded by the Korea government(MSIT) (No. RS-2023-00252434) and in part by the Chung-Ang University Research Grants in 2025.

\bibliographystyle{elsarticle-harv}
\bibliography{egbib}

\end{document}